# Clusters, Graphs, and Networks for Analysing Internet-Web Supported Communication within Virtual Community

Xavier Polanco

*Unité de Recherche et Innovation, Institut de l'Information Scientifique et Technique,*
*Centre National de la Recherche Scientifique, France*

**Abstract:** The proposal is to use clusters, graphs and networks as models in order to analyse the Web structure. Clusters, graphs and networks provide knowledge representation and organization. Clusters were generated by co-site analysis. The sample is a set of academic Web sites from the countries belonging to the European Union. These clusters are here revisited from the point of view of graph theory and social network analysis. This is a quantitative and structural analysis. In fact, the Internet is a computer network that connects people and organizations. Thus we may consider it to be a social network. The set of Web academic sites represents an empirical social network, and is viewed as a virtual community. The network structural properties are here analysed applying together cluster analysis, graph theory and social network analysis.

This is a work having taken place in the EICSTES project. EICSTES means *European Indicators, Cyberspace, and the Science-Technology-Economy System*. It is a research project supported by the Fifth Framework Program of R&D of the European Commission (IST-1999-20350.)

(7[th] International ISKO Conference, Granada, Spain, 10-13 July 2002. *Advances in Knowledge Organization*, Volume 8, Würzburg: ERGON Verlag, (ISBN 3-89913-247-5; ISSN 0938-5495), p. 364-371).

**1. Introduction**

We are concerned with Web structure analysis. Clusters, graphs, and networks procedures will be used to achieve it. The Internet is a computer network that connects people and organizations. As Garton et al. (1997) say "When a computer network connects people or organization, it is a social network." These authors argue the usefulness of a social network approach for the study of computer-mediated communication (CMC). This is our case here. The computer network is the Internet, and people or organizations are represented by a sample of 791 European Union academic Web sites. A community is called virtual community when it is a computer-supported social network in which communication among people is computer-mediated. Our proposal is to analyse the structure properties of the computer-mediated communication (CMC) within a virtual community. In this article, the proposition of using *clusters*, *graphs* and *networks* as models for analysing the Web structure will be examined.

Researchers from either information science or computer science converge to analyse the Web in terms of graph theory (Chakrabarti et al., 1999; Broder et al., 2000; Barabasi, 2001), and social network approach (Garton et al., 1997; Chakrabarti, 2000). We first clustered the Web network data (Polanco et al., 2001) and now we try here to aggregating clusters into a network. With regards to standard social network analysis this is an opposite approach. In social network analysis, the network is the prior given unit of analysis and cluster recognition corresponds to step of partitioning networks into subgroup components. In the case present we start from clusters and use the concepts of graph theory to recognize networks. Another difference in our approach is that we do not analyse the Web as a graph directly as it is usual in the Web studies.

**2. Web Data Sample**

The empirical reality is the Web, in our case a sample of 791 academic Web sites from the 15 countries belonging to the European Union, which are grouped into 37 clusters. The problem is the structural analysis of this empirical reality that is a large network in itself. Clusters are our starting blocks of analysis. Now the clusters will be decoded in terms of graph theory and social network analysis.

The Web is a valued directed graph whose nodes correspond to static pages and whose arcs correspond to hyperlinks between these pages. A *directed graph* consists of a set of nodes, denoted *V* and a set of arcs, denoted *E*. Each arc is an ordered pair of nodes (*i,j*) representing a directed connection from *i* to *j*. The *out-degree* of a node *i* is the number of links from *i* ($i,j_1$)...($i,j_k$), and the *in-degree* is the number of links to *i* ($j_1,i$)...($j_k,i$). From this reality, we have built a representation in which patterns that are hidden in the first reality they become uncovered, and constitute the building block of the analysis.

---

791 Web sites
5.819.674 Hyperlinks:
- (*i,j*) (*j,i*) = 5.308.204 out-links and in-links (91%)
- (*i,i*) = 511.470 self-links (9%)

12.595.809 pages

---

**Table 1:** Web Data Set. These data were collected in January 2001 by M. A. Boudourides and his co-workers at the Computer Technology Institute of Patras, Greece, as part of the project EICSTES.

Let us recall briefly how the sample of 791 academic Web site has been grouped into 37 clusters (see Polanco et al., 2001). The data matrix is a *N*-square matrix noted *D* where *N* is equal to the number of sites considered in the data set. The data matrix *D* recorded the number of hyperlinks between the *N* sites, in the diagonal the self-hyperlinks denoted (*i,i*), in the rows the directed hyperlinks denoted (*i,j*), (*j,i*) between the *N* Web sites. From this data matrix we may directly analyse the Web as a directed graph consisting of a set of nodes with directed arcs between pairs of nodes. For the study of the Web in directed graph terms, see (Broder. et al., 2000). Though our choice was to built another reality represented by a co-occurrence matrix. In this matrix patterns that are hidden in the data matrix become uncovered, and constitute the building block of the analysis. The hidden pattern is the co-occurrence of a pair of sites in the set of hyperlinks.

**3. Co-Site Analysis**

The approach that we adopted and we have called *co-site analysis* consists in recognizing couples of sites. Co-site is defined as the frequency with which two sites are co-associated together in the out-hyperlinks of a set of sites. Co-site analysis agrees with patterns of co-citation (Small, 1973 ; 1999) and co-word analysis (Callon et al., 1986). These approaches follow a general co-occurrence model. The difference is at the level of the object considered to co-occur. Thus it is possible to uncover a relationship among a pair of items (authors, words, or sites) that do not exist to first and directly approach.

Co-site is a relationship which is established by the hyperlinks of the other sites. The co-site frequency of two sites can be determined by computing lists of outgoing hyperlinks between the sites in the Web. Each of the two sites is located in the set of hyperlinks between sites in a given Web sample, and the number of sites at the origin of the hyperlinks defines the frequency of co-occurrence between the two sites. A new linking item is simply a new Web site that has hyperlinks with both sites. Co-site is the frequency with which two sites are associated together by the hyperlinks of the other sites.

In measuring co-site strength, we measure the degree of relationship or association between sites as perceived by the population of hyperlinking sites. These patterns can be changed over time because of the dependence on the hyperlinking sites. Such as vocabulary co-occurrences can change as subject domains evolve. The hyperlinking sites are those initiating the hyperlink that terminates at the sites receiving the hyperlink. Just as the distinction between citing and cited in co-citation analysis. Co-site patterns change as the interest and information exchange patterns of the considered field in the Web change.

When two sites are frequently associated by the hyperlinks of the other sites, they are also necessarily frequently pointed by the hyperlinks of the other sites individually as well. Frequently pointed sites maybe represent the key Web sites in a given domain. Thus co-site patterns can be used to map out the relationships between these key Web sites. This allows a way of modelling the communication structure of a particular set of Web sites. Changes in the co-sites patterns, when they are considered over a period of time, may provide clues to understanding the mechanism of a Web domain development.

Network of co-sites can be generated for specific sectors of the Web, and then submitted to clustering. Clusters of Web co-sites provide a new way to study the Web structure. In general a clustering method attempts to reorganize some entities into relatively homogeneous groups. Thus clusters represent groups of highly similar Web sites. Similarity may be based on the degree of relationship or association that exists between entities. We used an association coefficient to measure degree of similarity between sites described by binary data, 1 refers to the presence of a variable (co-occurrence) and 0 to its absence. The clustering method that we used was a hierarchical agglomerative method following a single linkage rule. From this result, a set of clusters, the task is now to produce a network of clusters, a cluster network following graph theory.

A cluster consists of five sets of information: [1] a set of components, [2] a set of internal associations between pairs of components (or co-sites), [3] a set of external associations between pairs of components belonging to different clusters, [4] a set of values attached to the associations, and finally [5] a set of sites clustered. All the associations or relationships have a strength value. These relations are based on the co-occurrence. Each cluster represents itself a valued graph. In the next section we deal only with inter-cluster relations in graph-theoretic terms.

## 4. A Network of Clusters

Following Degenne and Fossé (2001), Wasserman and Faust (1999), we shall revisit mainly the relations between clusters, called inter-cluster relations, according to graph-theoretic concepts. Let each cluster be a node, *n*, thus *n* is equal to 37 nodes. An illustration of this co-site network is shown in Figure 1. Each numbered box in the network diagram represents a cluster of co-sites. The inter-cluster relations are both directional and valued relations. Thus the network of clusters would be represent by a *valued directed graph*. For simplifying, the graph that is displayed in Figure 1, however, is this that conforms to the following restriction. A binary directed graph, in which the strength values as well as the amount of relations between two clusters are not considered. This restriction forms a natural starting point for modelling a network of clusters. It introduces a minimum amount of arbitrary structure, whilst still allowing meaningful questions to be asked of the network as whole.

### 4.1 Directed Graph

The inter-cluster relations are directional relations. A relation is directional if the relation is oriented from one node to another. Directed relations between pairs of nodes are represented as lines in which the directions of the relations are specified by the arrowheads. These oriented lines are called *arcs*. An arc is an ordered pair of nodes reflecting the direction of the relation between two nodes.

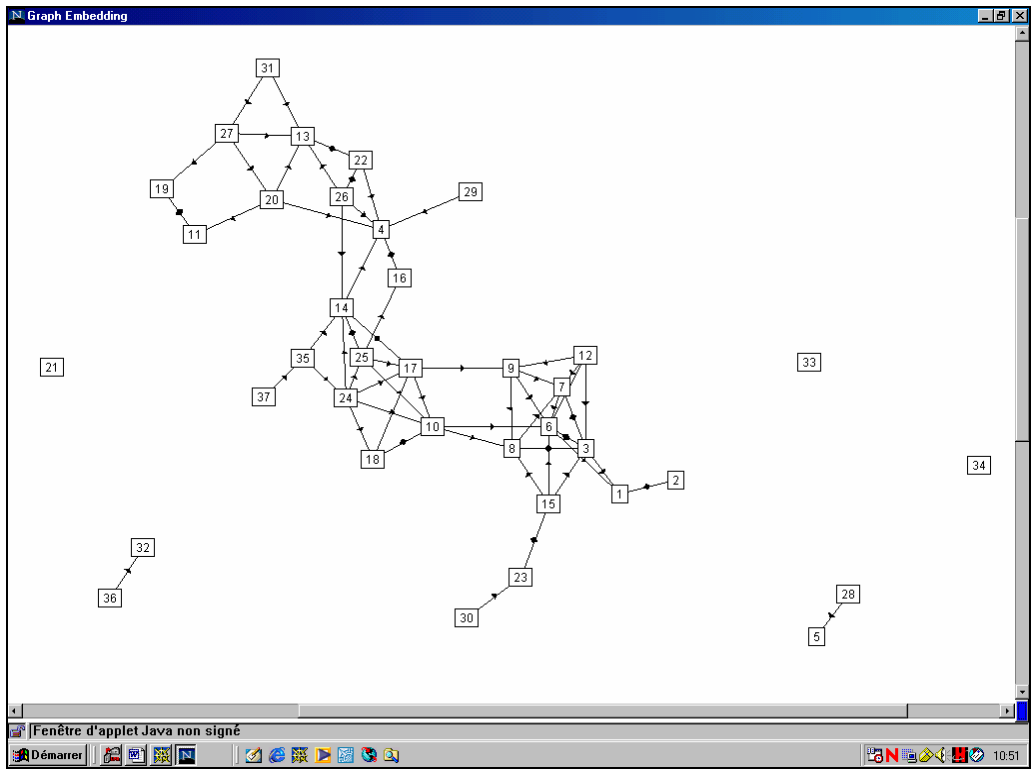

**Figure 1:** The directed graph representing the 37 clusters and their interrelations. Each numbered box represents a cluster of co-sites. The directions of the inter-cluster relations are specified by the arrowheads.

The clusters can be considered as nodes of a directed valued graph. A directed graph, $G_d(N,L)$ consists of two sets of information: a set of nodes $N = \{n_1, n_2, …, n_N\}$, and a set of arcs, $L = (l_1, l_2, …, l_L)$. Since each arc is an ordered pair of nodes, there are $N(N-1)$ possible arcs in $L$. Each arc is an ordered pair of distinct nodes, $l_k = (n_i, n_j)$. The arc $(n_i, n_j)$ is directed from $n_i$, the origin or sender node, to $n_j$, the terminus or receiver node. A node is incident with an arc if the node is in the ordered pair of nodes defining the arc. The nodes $n_i$ and $n_j$ are incident with the arc $l_k = (n_i, n_j)$. Since an arc is an ordered pair of nodes, we can distinguish the first from the second node in the pair, and we must consider if a given node is *sender* or *receiver* in the ordered pair defining the arc. Formally, node $n_i$ is *adjacent to* node $n_j$ if $(n_i, n_j) \in L$, and node $n_j$ is *adjacent from* node $n_i$ if $(n_i, n_j) \in L$.

In Figure 1, the clusters are represented as numbered boxes and the arcs are represented as directed arrows. The arc $(n_i, n_j)$ is represented by an arrow from the point representing $n_i$ to the point representing $n_j$. For example, if cluster $i$ has a relation with cluster $j$ there is an arc originating at $i$ and terminating at $j$. If cluster $j$ returned the tie, there is another arc, this one originating at $j$ and terminating at $i$.

In a directed graph, or digraph, three types of relations occur between the $N(N-1)/2$ pairs of nodes: [1] *mutual*, with both nodes directing relations toward each other, shown by two-headed arrows ($n_i \leftrightarrow n_j$); [2] *asymmetric*, in which one node directs a relations toward another that is not reciprocated ($n_i \rightarrow n_j$); [3] *null*, in which no relation in either direction exists between a pair of nodes. Figure 1 shows all three types of relations. These patterns can be observed in Figure 1. One might also observe that the graph is far from complete. A graph is said to be complete if all $N(N-1)$ possible relations between the set of $N$ nodes are present.

**4.2 Valued Graph**

As mentioned above, the network of clusters consists of valued relations in which the strength of each relation is recorded. In the case of valued relations, valued graphs are the appropriate graph-theoretic representation. A *valued graph* or a *valued directed graph* is a graph (or digraph) in which each line (or arc) carries a value. A valued graph, $G(N,L,V)$, consists of three sets of information: a set of nodes (or vertex, or points), $N = \{n_1, n_2, \ldots, n_N\}$, a set of lines (or arcs or edges), $L = \{l_1, l_2, \ldots, l_L\}$, and a set of values, $V = \{v_1, v_2, \ldots, v_V\}$. Associated with each line (in a graph) or each arc (in a digraph) is a value from the set of real numbers. In our case, the values result from the formula $E_{(ij)} = C_{(ij)}^2 / C_{(i)} \times C_{(j)}$, where $C_{(ij)}$ denotes the number of co-occurrences of a pair of sites $(i,j)$ as receivers in the set of the hyperlinks of the other sites; $C_{(i)}$ and $C_{(j)}$ the number of occurrences of sites $(i)$ and $(j)$ as receivers in the set of the hyperlinks of the other sites.

In a valued graph the relation between node $n_i$ and node $n_j$ is identical to the relation between node $n_j$ and node $ni$, $l_k = (n_i, n_j) = (n_j, n_i)$, and thus there is only a single value, $v_k$, for each unordered pair of nodes. It is the case of intra-cluster structure. It is not the case of the relations among the clusters, or inter-cluster relations, which obey an ordered pair of nodes as effects of the clustering method. The order is imposed at the level of clusters by the order of creation of clusters. The relation itself is an undirected valued relation having a single value, $v_k$, for each pair of nodes each one belongs to two different clusters.

**4.3 Valued Directed Graph**

The network of clusters should be represented as a *valued directed graph* in which each cluster is a *valued undirected graph*. A *valued directed graph* or a *weighted digraph* represents a directional valued relation, such as the amount of links from each Web site to each other Web site. Site $i$ may address a different amount of links to site *j* than site *j* address to site *i*. In a valued directed graph, the arc from node $n_i$ to node $n_j$ is not the same as the arc from node $n_j$ to node $n_i$, $l_k = (n_i,n_j) \neq l_m = (n_j,n_i)$, and thus there are two distinct values, one for each possible arc for the ordered pair of nodes. In general, for $l_k = (n_i,n_j)$ and $l_m = (n_j,n_i)$, $v_k$ does not necessarily equal $v_m$.

The inter-cluster relations among the 37 clusters can be analysed in terms of a valued directed graph. The values that we can consider are the strength values of each relation between a pair of sites, and the out-degree and in-degree values for each cluster. For simplifying, the graph that is displayed in Figure 1, however, is a binary directed graph, in which the strength values as well as the amount of relations between two clusters are not considered. This graph is built from the *N* square matrix of out-degree and in-degree binary values for each cluster where *N* is equal to the number of clusters, that is, *N* = 37. In-degrees and out-degrees are useful measurements in particular for our type of networks and relations in which an information exchange occurs.

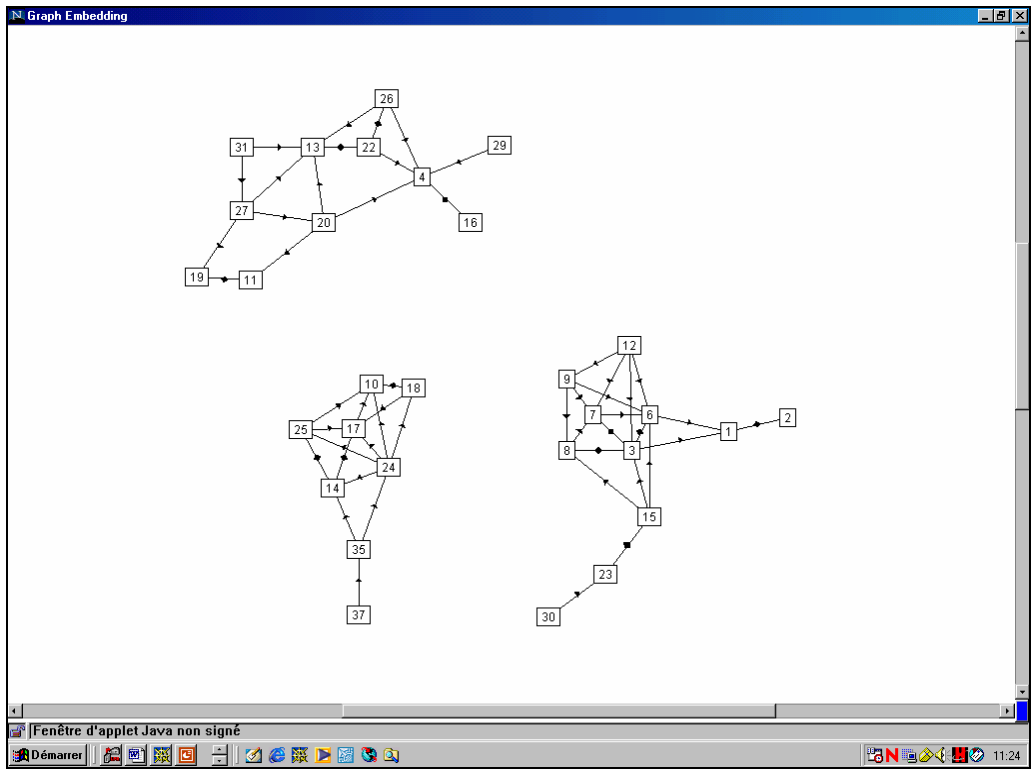

**Figure 2:** The sub-graphs in which the network may be partitioning. The arcs connecting them have been remote. The removed arcs are the following (see Figure 1): (17→9), (10→6), (10→8), and (14→4), (26→14), (25→16). All the arcs represent single directional relations between the clusters.

In a graph, the degree of a node, denoted by $d(n_i)$, is the number of nodes adjacent to it, equivalently, the number of lines incident with it. In a digraph, a node can be either *adjacent to*, or *adjacent from* another node, depending on the direction of the arc. It is interesting to consider these cases separately. Since one quantifies the tendency of nodes to be senders; the other quantifies the tendency to receive. A node with degree equal to 0 is called an isolate.

− *Nodal out-degree*. The out-degree of a node, $d_O(n_i)$, is the number of nodes adjacent from $n_i$. The out-degree of node $n_i$ is equal to the number of arcs of the form $l_k = (n_i, n_j)$, for all $l_k \in L$, and all $n_j \in N$. The out-degree value measures the number of arcs originating with any node $n_i$.

- *Nodal in-degree.* The in-degree of a node, $d_I(n_i)$, is the number of nodes that are adjacent to $n_i$. The in-degree of node $n_i$ is equal to the number of arcs of the form $l_k = (n_j, n_i)$, for all $l_k \in L$, and all $n_j \in N$. The in-degree value measures the number of arcs terminating at any $n_i$.

In our example of inter-clusters relationships, clusters with high out-degree can be recognized as *senders*, and clusters with high in-degree as heavy *receivers* in the exchange of information. There are also isolated clusters. Note these clusters are each one a set of co-sites. This is a way to discover the isolated subset of sites within the network. Isolate means here to be close inside of a cluster without connections with any other cluster. Also pairs of clusters are distinguished; they are only related together without relations with the other clusters. Thus they are together isolated from the others in the network.

In terms of the in-degrees and out-degrees of the nodes in a directed graph, we can distinguish four different kinds of nodes (Wasserman and Faust, 1999). [1] The node is an isolate; [2] the node only has arcs originating from it; [3] the node only has arcs terminating at it; [4] the node has arcs both to and from it. All these cases can be observed in our example as shown in Figures 1 and 2.

**5. A Network of Networks**

Finally, we may consider the overall structure of the network took as a whole. From this point of view, three sub-networks can be recognized in Figure 1. These three sub-networks are shown in a separate manner in Figure 2. The arcs connecting them have been remote. This is a way of analysing the whole network. Furthermore, the information that each cluster represents allows knowing the academic sites that are together in each sub-network. The example is chosen mainly to illustrate how the whole network can be interpreted in a real situation. The cities and countries in which they are located may also be considered. We also expect to observe changes in the network patterns, when they are considered over a period of time. This approach may provide clues to understanding the mechanism of a Web domain development.

In this framework, many different indices can be computed from matrices to measure structural characteristics for both individual actors and entire networks. This issue is a important subject that is out of the scope of this article. We then stop here.

**6. Conclusions**

Let us recall that our research proposal deals with using *clusters*, *graphs* and *networks* as models for analysing the Web structure. This is in progress. In this article, we have limited to use graph theoretic concepts for analysing clusters of co-sites. The issue that remains to be considered in the framework of pattern recognition and exploratory data analysis is the clustering methods based in graph theory (see Hubert, 1974; Dubes and Jain, 1980; Theodoridis and Koutroumbas, 1999).

It appears that an interpretation of the significance of the clusters of co-sites must rely on the notion of similarity, and on the association or co-occurrence of contents. Co-site clusters maybe correspond to significant intellectual connections within the Web field in consideration. This suggests extending the co-sites analyse into Web content analysis.

Another area for the application of co-site analysis is in the study of the structure of science in the Web. We are thinking in the co-site analysis restated as we began to do here in the framework of graph theory and social network analysis. The pattern of relations among key R&D sites establishes a structure for the scientific specialty which may then be observed to change through time. Through the study of these changing structures, co-site analysis becomes a tool for analysing and monitoring the development of scientific fields on the Web, and for assessing the degree of interrelationship among specialties in the Web context, as well as co-citation and co-word analysis in the context of bibliographic databases since a long time in the study of science.

**References**

Barabasi, A.-L. (2001) The Physics of the Web. *Physics World*, vol. 14 (7) pp. 1-12. [http://www.physicsweb.org/]

Broder, A., R. Kumar, F. Maghoul, P. Raghavan, S. Rajagopalan, R. Stata, A. Tomkins, J. Wiener. (2000) Graph structure in the Web. *Computer Networks* vol. 33, pp. 309-320.

Callon, M., J. Law, A. Rip. (1986) *Mapping the Dynamics of Science and Techn*ology. London, The Macmillan Press.

Chakrabarti, S. (2000) Data Mining for Hypertext: A Tutorial Survey. *SIGKDD Explorations*, vol. 1 (2), pp. 1-11.

Chakrabarti S., B. E. Dom, S. Ravi Kumar, P. Raghavan, S. Rajagopalan, A. Tomkins, D. Gibson, J. Kleinberg. (1999) Mining the Web's Link Structure. *Computer*, pp. 60-67.

Degenne A., M. Forsé. (2001) *Les réseaux sociaux*. Paris, Armand Colin.

Dubes, R., A. Jain. (1980) Clustering Methodologies in Exploratory Data Analysis. *Advances in Computers*, vol. 19, pp. 113-228.

Garton, L., C. Haythornthwaite, B. Wellman (1997) Studying Online Networks. *JCMC* vol. 3 (1) June 1997. [http://www.ascusc.org/jcmc/vol3/issue1/garton.html]

Hubert, L. J. (1974) Some Applications of Graph Theory to Clustering. *Psychometrika*, vol. 39 (3), pp. 283-309.

Polanco, X., M. A. Boudourides, D. Besagni, I. Roche (2001) Clustering and Mapping European University Web Sites Sample for Displaying Associations and Visualizing Networks, in ETK-NTTS 2001, *New Techniques and Technologies for Statistics Exchange of Technology and Know-how*, Pre-proceedings of the Conference, Hersonissos, Crete, 18-22 June 2001, vol. 2, pp. 941-944.

Theodoridis, S., K. Koutroumbas (1999) *Pattern Recognition*. San Diego, CA., Academic Press.

Small, H. (1973) Co-citation in the Scientific Literature: A New Measure of the Relationship between Two Documents. *Journal of the American Society for Information Science*, vol. 24 (July-Aug.), pp. 265-269.

Small, H. (1999) Visualizing Science by Citation Mapping. *Journal of the American Society for Information Science*, vol. 50 (9), pp. 799-813.

Wasserman S., K. Faust (1999) *Social Network Analysis. Methods and Applications*. Cambridge, Cambridge University Press.